\documentclass[letterpaper, 10pt, conference]{ieeeconf}
\IEEEoverridecommandlockouts	% to override locked commands

\usepackage{multirow}
\usepackage{lipsum}
\usepackage{amsmath}
\usepackage{amssymb}
\usepackage{textcomp}
\usepackage{gensymb}
\usepackage[ruled,vlined]{algorithm2e}
\usepackage{graphicx}
\usepackage{euscript}
\usepackage{ctable}
\graphicspath{{./Figures/}}
\usepackage[nonumberlist,acronym]{glossaries}
\usepackage[hidelinks]{hyperref}

\newacronym{ROS}{ROS}{Robot Operating System}
\newacronym{LiDAR}{LiDAR}{Light Detection and Ranging}
\newacronym{ToF}{ToF}{Time-of-Flight}
\newacronym{RMSE}{RMSE}{Root Mean Square Error}

\begin{document}

\title{Deterministic Guided LiDAR \\Depth Map Completion}

%\author{\IEEEauthorblockN{Bryan Krauss, Gregory Schroeder, Marko Gustke, and Ahmed Hussein}
%	\IEEEauthorblockA{IAV GmbH, Intelligent Systems Functions Department, Berlin, Germany\\
%		Emails: bryan@krauss-tmk.com, gregory.schroeder@iav.de, marko.gustke@iav.de, ahmed.hussein@ieee.org}
%}

\author{Bryan Krauss, Gregory Schroeder, Marko Gustke, and Ahmed Hussein%
	\thanks{IAV GmbH, Intelligent Systems Functions Department, Berlin, Germany \newline
		{\tt\small \{bryan.krauss, gregory.schroeder, marko.gustke\}@iav.de, ahmed.hussein@ieee.org}}%
}

\maketitle
%\raggedbottom
\thispagestyle{empty}
\pagestyle{empty}

%%%%%%%%%%%%%%%%%%%%%%%%%%%%%%%%%%%%%%%%%%%%%%%%%%%%%%%%%%%%%%%%%%

\begin{abstract}

Accurate dense depth estimation is crucial for autonomous vehicles to analyze their environment. This paper presents a non-deep learning-based approach to densify a sparse LiDAR-based depth map using a guidance RGB image. To achieve this goal the RGB image is at first cleared from most of the camera-LiDAR misalignment artifacts. Afterward, it is over segmented and a plane for each superpixel is approximated. In the case a superpixel is not well represented by a plane, a plane is approximated for a convex hull of the most inlier. Finally, the pinhole camera model is used for the interpolation process and the remaining areas are interpolated. The evaluation of this work is executed using the KITTI depth completion benchmark, which validates the proposed work and shows that it outperforms the state-of-the-art non-deep learning-based methods, in addition to several deep learning-based methods.

\end{abstract}

%%%%%%%%%%%%%%%%%%%%%%%%%%%%%%%%%%%%%%%%%%%%%%%%%%%%%%%%%%%%%%%%%%

\section{Introduction}
\label{sec:introduction}

A dense and accurate perception of the 3D environment is crucial in several research fields; such as autonomous driving. Common sensors for depth estimation are monocular and stereo cameras, as well as~\gls{LiDAR} sensors. While images from cameras estimate the depth with inaccuracy, a~\gls{LiDAR} provides quite accurate depth measurements and is therefore used more often. 

\begin{figure}[ht]
	\begin{center}
		\includegraphics[width=0.48\textwidth]{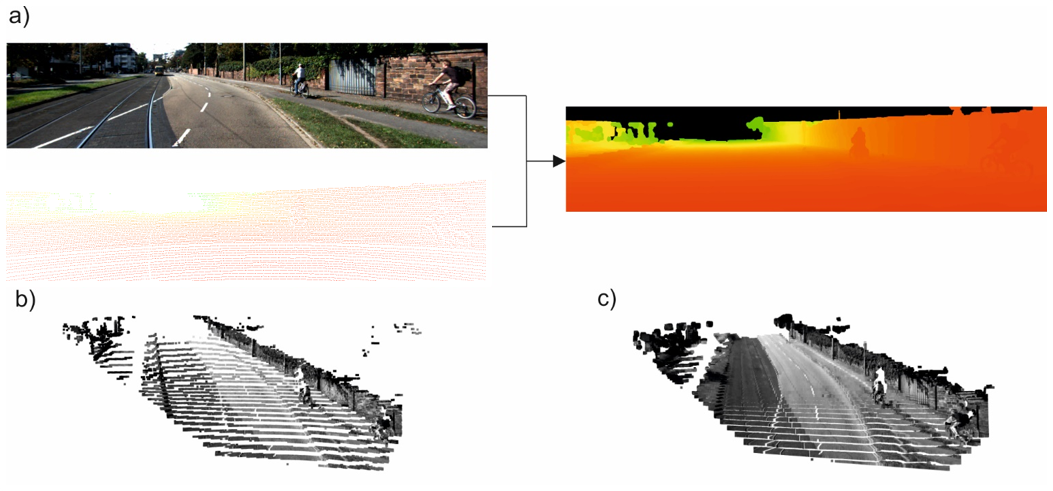}
		\caption{a) Fusion of RGB-image and sparse depth map to obtain a denser depth map; b) Colorized sparse~\gls{LiDAR} pointcloud; c) Colorized pointcloud from a dense depth map}
		\label{fig:intro}
	\end{center}
\end{figure}

However,~\gls{LiDAR} provides only sparse depth measurements, due to hardware limitations. The sparsity poses difficulties for perception algorithms to utilize~\gls{LiDAR} depth measurements. Therefore, modern research in academia and industry attempts to overcome these limitations by using the dense color information of a camera image to densify the~\gls{LiDAR} measurements, as shown in Figure~\ref{fig:intro}. 

Next to the improvement of visual-based algorithms by providing dense RGB-D data, the fusion also enables the reduction of~\gls{LiDAR} layers, which yields enormous cost reduction. Deep learning methods have gained popularity to tackle this problem, however, they often require expensive hardware, big data, and long training time. Furthermore, since they are non-deterministic, a certification for deep learning approaches in series production is problematic.

Accordingly, the research question in this paper revolves around the possibility to fuse color information from a camera and~\gls{LiDAR} data in a deterministic way, thus the sparse~\gls{LiDAR} depth maps are densified. The hypothesis is to be validated over the KITTI dataset and compared to other approaches from the literature. The contribution of this paper is to propose a deterministic algorithm to densify~\gls{LiDAR} depth map, guided by colored images, in near real-time. However, because the scene contains complex structures that can not be approximated by planes, this work needs a fusion algorithm to interpolate the remaining areas.

The remainder of this paper is organized as follows; Section~\ref{sec:RelatedWork} reviews the related work from the literature. Section~\ref{sec:Methodology} describes the proposed approach to densify the sparse depth maps. This is followed by Section~\ref{sec:ExperimentalWork}, which gives insights into the used setup, evaluation metrics, and the evaluation scenario set. In Section~\ref{sec:ResultsAndDiscussion}, the obtained results are presented in quantitative and qualitative analysis. Finally, Section~\ref{sec:conclusion} summarizes the paper and mentions recommendations for future research.

%%%%%%%%%%%%%%%%%%%%%%%%%%%%%%%%%%%%%%%%%%%%%%%%%%%%%%%%%%%%%%%%%%

\section{Related Work}
\label{sec:RelatedWork}

Generally, the field of depth completion is divided into two categories: guided and non-guided depth completion. On the one hand, non-guided approaches are solely based on depth information and does not use additional sensor information. On the other hand, guided approaches are usually based on the depth information itself and a guidance image. 

A variety of related approaches exist in the field of~\gls{ToF} super-resolution and~\gls{ToF} depth map enhancement. Although there are differences between mechanical laser scanners and~\gls{ToF} cameras, in terms of environment conditions, depth ranges, number of points and their distribution; the approaches towards upsampling and depth map completion might be applied to both sensor technologies. 

\subsection{Non-guided Depth Completion}
\label{sub:depComp}

Regarding non-guided depth completion, the variety of approaches shrinks. Zhou~\cite{zhou2018depth} proposed a sole depth map based edge-guided upsampling method to tackle the~\gls{ToF} super-resolution task, which creates a high-resolution edge-map using sparse coding and finally uses a modified joint bilateral filter (replaced color term with step function regarding if the point is on the same side of the edge) to obtain the final high-resolution depth map. 

Premebida~\cite{premebida2016high} proposed an approach to complete~\gls{LiDAR} depth maps solely based on~\gls{LiDAR} data by filling all unknown depth values with its nearest neighbors and afterward applying a modified bilateral filter, which uses DBSCAN to detect discontinuities between points.

Uhrig~\cite{uhrig2017sparsity} created the KITTI depth completion benchmark for the evaluation of their algorithm, which created an end-to-end CNN consisting of sparse convolution layers (using~\gls{LiDAR} only). Ku~\cite{ku2018defense} use a set of morphological operations to interpolate the missing depth values. Hekmatian~\cite{hekmatian2019conf} proposed an end-to-end network, which predicts the densified depth map as well as an error-map to generate high confidence values.

\subsection{Guided Depth Completion}
\label{sub:ToF}

The pioneers in the field of~\gls{ToF} guided upsampling formulated this task usually as an optimization problem with differing regularization term, as presented in~\cite{diebel2005application, yang2007spatial, park2011high, ferstl2013image}. However, a more recent approach is introduced by~\cite{barron2016fast}, which formulated the upsampling as an optimization problem in the bilateral space.

In~\cite{he2012guided}, authors introduced the guided filter, which assumes a linear model between local depths and color information. Furthermore, they introduced the fast guided filter in~\cite{he2015fast}, which builds a linear model on a subsampled version of the depth map and guidance image. Thereupon, in~\cite{konno2015intensity}, authors used this idea as an initial step and operates in the residual domain. Additionally, \cite{bapat2019domain}, authors used the properties of the edge-preserving domain transform filter to upsample~\gls{ToF} depth maps. 

Another class of methods is based on~\cite{kopf2007joint}, which obtained a high-resolution depth map by interpolating missing depths by a weighted sum of the spatial and color distance of supporting pixel with known depth, termed as joint bilateral upsampling. Then, in~\cite{lo2017edge}, authors used a tentative bicubic upsampled depth map to iteratively refine it using a modified joint trilateral guided filter, which uses local gradient information. 

Zhang~\cite{zhang2016depth} tackled the enhancement of depth maps by at first filling large holes using a patch-based inpainting method and afterward filtering the depth map using color, spatial, and depth difference (joint trilateral guided filter). Furthermore, Tian~\cite{tian2016color} enhanced the~\gls{ToF} depth map by aligning edge information from the depth map and the color image and afterward use drift vectors and a maximal bilateral filter to generate a final depth map. Chen~\cite{chen2016bilateral} proposed the bilateral guided upsampling approach which finds application in image upsampling. It uses the 3D bilateral grid with each cell representing an affine transformation between the guidance high resolution and the low-resolution image to obtain the final upsampled high-resolution image.

Regarding the fusion of~\gls{LiDAR} and camera in the KITTI depth completion benchmark~\cite{uhrig2017sparsity}, Yao~\cite{yao2020discontinuous} is the only published and peer-reviewed paper listed in the KITTI depth completion benchmark, which is non-deep learning-based. This approach uses a binary anisotropic diffusion tensor to interpolate missing depth values. Furthermore, there are several deep learning-based approaches. Ma~\cite{ma2019self} proposed an encoder-decoder network where the encoding part is based on the ResNet-34 to tackle the depth completion task. Furthermore, the authors proposed a self-supervised framework, which uses the transformation of a previous frame into the current frame to calculate the prediction error. 

Dimitrievski~\cite{dimitrievski2018learning} proposed a framework, which applies Contraharmonic Mean Filter layers to approximate morphological operations, which then servers as input for a U-Net architecture. Tang~\cite{tang2020learning} uses an encoder-decoder framework, which generates content-dependent and spatially-variant kernels. Qiu~\cite{qiu2019deeplidar} proposed an encoder-decoder structure, which fuses a depth map generated using RGB and sparse~\gls{LiDAR} data with a predicted surface normal map and confidence mask to generate a final dense depth map. Chen~\cite{chen2019learning} propose a framework, which on the one hand applies 2D convolutions on the RGB image and on the other hand applies continuous convolutions on the 3D points and fuses both obtained features.

Due to the success of DeepLiDAR~\cite{qiu2019deeplidar} in the depth completion benchmark, and the successful idea to approximate the 3D scene by a piece-wise planar scenario in the field of computer vision, the proposed work develops a deterministic method that uses this geometric information for an interpolation process.

%%%%%%%%%%%%%%%%%%%%%%%%%%%%%%%%%%%%%%%%%%%%%%%%%%%%%%%%%%%%%%%%%%
\section{Proposed Approach}
\label{sec:Methodology}

Assume a depth map $D \in \mathbb{Z}^2$ with its coordinates $(u,v) \in D$. Image vectors are denoted by $\vec{x} = [u,v]^T$, whereas object vectors in $\mathbb{R}^3$ are denoted by $\vec{\mathcal{X}} = [X,Y,Z]^T$. Homogeneous vectors are denoted by a superscript e.g. $\vec{x}^\prime$. Furthermore, assume a proper camera projection matrix $P = [M | p_4]$ with $p_4$ as the $4$-th column of $P$ and thus $M \in \mathbb{R}^{3\times3}$. $\tau$ represents thresholds which has to be determined experimentally. At first, this section will give an overview of artifacts, which influences the quality of the proposed work and restricts assumptions.

\subsection{Artifacts}
\label{sub:Artifacts}

\begin{figure}[ht]
	\begin{center}
		\includegraphics[width=0.48\textwidth]{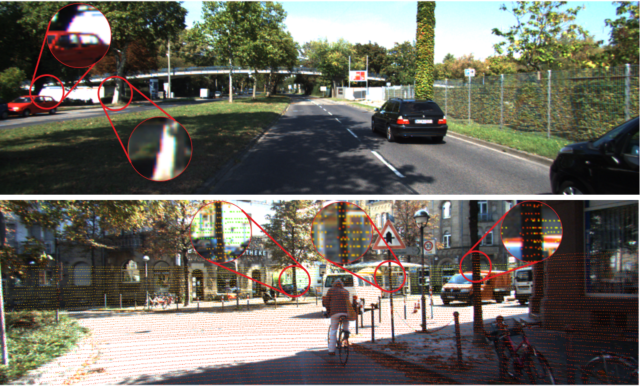}
		\caption{(Top) Color fringing; (Bottom) Misalignment artifact}
		\label{fig:artifacts}
	\end{center}
\end{figure}

In this section, artifacts that influence the quality of guided~\gls{LiDAR} depth completion are shown. Related work in this field does not point out these artifacts. The next points give a brief overview of these artifacts:

\begin{itemize}
	\item \textbf{Image artifacts:} next to common image artifacts (blurring, distortion etc.) depth completion suffers from color fringing (top image in Figure~\ref{fig:artifacts}) which emerges in RGB images at bright-dark transitions. Thus, object boundaries are not clearly visible. Without the raw-image format, this artifact can not be mitigated.
	
	\item \textbf{Reflective surfaces:}~\gls{LiDAR} usually have a weakness against high reflective surfaces and thus creating false depth values.
	
	\item \textbf{Discretization error:} reconstructing a pointcloud from the depth map is affected by the discretization error due to the loss of information when projecting points from $\mathbb{R}^3$ into the integer depth map $D \in \mathbb{Z}^2$.
	
	\item \textbf{Misalignment artifact:} camera and~\gls{LiDAR} can not be installed at the same position, thus have a different FOV. This results in projecting background points into the foreground shown at the bottom of Figure~\ref{fig:artifacts}. Because background points are projected into the same vertical layers and thus in a close neighborhood $\EuScript{N}$ of the foreground points, applying Equation~(\ref{eq:preproc}) to each~\gls{LiDAR} measurement $x_{lid} \in D$ mitigates the influence of the misalignment artifact.
	
\end{itemize}

	\begin{equation}
		\label{eq:preproc}
		D(u_{\EuScript{N}}, v_{\EuScript{N}}) = 	
		\begin{cases}
			0 & \!\begin{aligned}%[b]
				& D(u_{\EuScript{N}}, v_{\EuScript{N}}) \geq  \\
				& D(u_{lid}, v_{lid}) \cdot \tau_{\EuScript{N}}
			\end{aligned} \\%[1ex]
			D(u_{\EuScript{N}}, v_{\EuScript{N}}) & \text{else}
		\end{cases},
		\forall x_{lid} \in D
	\end{equation}

\begin{itemize}
	
	\item \textbf{Different acquisition times:} when the~\gls{LiDAR} moves each measurement in azimuth direction has a different origin and thus are not projected correctly (if not transformed into the same frame). However, both sensors do not only have to be transformed into the same origin but also need to be transformed into the origin when the camera acquired the scene. Otherwise, the acquired image information and depth measurements do not align. The more dynamic the scene is, the more different the acquired scene information is.
	
\end{itemize}

As most artifacts can hardly be removed in postprocessing, sensor- and setup-quality are essential. Utilizing RANSAC for plane approximation (see \ref{sub:plane}) and an additional validity check (see \ref{sub:Validity}) limits their impact though.

\subsection{Semantic Context}
\label{sub:SemCont}

Considering the entire scene at once is too complex and would not allow to approximate a piece-wise planar scene. Thus, the image is partitioned into multiple regions. Using a fix-sized grid in $u,v$-direction would disrupt any semantic information. A more useful approach is to oversegment the image into multiple superpixel which share the same color and thus usually belonging to the same object. Due to its runtime advantages, Simple Linear Iterative Clustering (SLIC) \cite{slic} is used as segmentation method. This results in a segmented RGB-Image $\Omega$ where each superpixel $S_i$ contains multiple associated~\gls{LiDAR} measurements $x_{lid}$ and their corresponding depth. A plane $\Pi \subset \mathbb{R}^3$ is then approximated for each superpixel and used to interpolate all unknown pixel.

Because superpixel can violate object boundaries and to avoid planes approximated using a small number of points (potentially unreliable), only superpixel are considered with a minimum number of~\gls{LiDAR} measurements $\tau_m$ and only superpixel which contains measurements from at least two different horizontal and two different vertical~\gls{LiDAR} scans (which as well ensures the non-collinearity constraint). 

Usually, $S_i$ contains~\gls{LiDAR} measurements which are well represented by $\Pi$. But the case can occur, that $S_i$ either describes a too complex (to approximate it well by $\Pi$) part of an object or multiple objects are contained in $S_i$. This is caused by different objects with same color, shadow/blending or poor segmentation. In this areas a convex hull $C$ is determined using the image points, which are well represented by $\Pi$. A plane is afterwards only applied to this convex hull.

\subsection{Plane Approximation}
\label{sub:plane}

Assume a set of known~\gls{LiDAR} measurements $K_i \subset S_i$. Because the $(u,v,Z)$-space is not suitable for plane approximation, all $\vec{x}^\prime_{lid} \in K_i$ are back-projected to $\mathbb{P}^3$ using equation~\ref{eq:backproj1} and afterwards converted to $\mathbb{R}^3$.

\begin{equation}
\label{eq:backproj1}
	\vec{\mathcal{X}^\prime}_{lid}(Z) = \begin{bmatrix} M^{-1}(Z \vec{x}^\prime_{lid} - p_4) \\ 1 \end{bmatrix}, \quad \forall \vec{x}^\prime_{lid} \in K_i
\end{equation}

This results in a pointcloud corresponding to $S_i$. Afterwards $\Pi$ is determined. Regression methods such as regularized regression and ordinary least squares suffers from the lack of plane information. Thus, a plane which minimizes the orthogonal loss $E_{TLS}$ to the vectors $\vec{\mathcal{X}_i}$ is sought. This plane is found by minimizing equation \ref{eq:ETLS}.

\begin{equation}
	\label{eq:ETLS}
	\min_{||\vec{n}||_2= 1} E_{TLS}(\vec{n}) = \min_{||\vec{n}||_2= 1} ||A \vec{n}||_2
\end{equation}

This minimization problem has a closed form solution by solving equation \ref{eq:homogen}

\begin{equation}
\label{eq:homogen}
	A \vec{n} = 0
\end{equation}

With 

\begin{equation}
	A = \begin{bmatrix}
		X_0 & Y_0 & Z_0 \\
		X_1 & Y_1 & Z_1 \\
		\vdots & \vdots & \vdots \\
		X_n & Y_n & Z_n\\
	\end{bmatrix}, \quad \vec{n} = [\Pi_0, \; \Pi_1, \; \Pi_2]^T
\end{equation}

with $n = |S_i|$. This can be solved by decomposing the matrix $A$ using the singular value decomposition (equation \ref{eq:SVD}).

\begin{equation}
	\label{eq:SVD}
	A = U \Sigma V^T
\end{equation}

The parameter vector $\vec{n}$ which minimizes the orthogonal loss, is the right singular vector corresponding to the smallest singular value. Given the normal $\vec{n}$, $\Pi_3$ need to be determined. This can be done using equation~\ref{eq:plane3}

\begin{equation}
	\label{eq:plane3}
	\Pi_3 = -\Pi_0 \overline{X} - \Pi_1 \overline{Y} - \Pi_2 \overline{Z}
\end{equation}

with $\overline{X}$ as the mean of all $X$-coordinates. 

\subsection{Interpolation Using Planes}
\label{sub:interpol}
Given the depth, a ray $\vec{l}(\vec{x}^\prime)$ joining the pixel $\vec{x}^\prime$ and the camera center can be used to determine a unique Cartesian coordinate as shown in Equation~(\ref{eq:backproj1}).

However, this process can be used to interpolate unknown depths. Assume a set $U_i \subset S_i$ consisting of homogeneous image vectors $\vec{x^\prime_k} \in U_i$ with unknown depths. The sought unique Cartesian coordinate for an unknown vector, is the intersection of $\Pi$ with the ray $\vec{l}(\vec{x}^\prime_k)$. This intersection $\vec{\mathcal{X}}_{int}$ can be calculated using equation \ref{eq:vecIntersection}

\begin{equation}
\label{eq:vecIntersection}
	\vec{\mathcal{X}}_{int} = w + s \cdot \vec{l}(\vec{x}_k^\prime) + p_0 ,\quad \forall \vec{x}_k^\prime \in U_i
\end{equation}

\noindent
where 

\begin{equation}
	s = \frac{-\vec{n} \cdot w}{\vec{n} \cdot \vec{l}(\vec{x}_k^\prime)} \quad, \quad w = l_0 - p_0 \quad and \quad \vec{l}(\vec{x}_k^\prime) = M^{-1}\vec{x}^\prime_k 
\end{equation}

\noindent
where $p_0 \in \Pi$ and $l_0 \in \vec{l}(\vec{x}^\prime_k)$. 

\begin{figure}[ht]
	\begin{center}
		\includegraphics[width=0.48\textwidth]{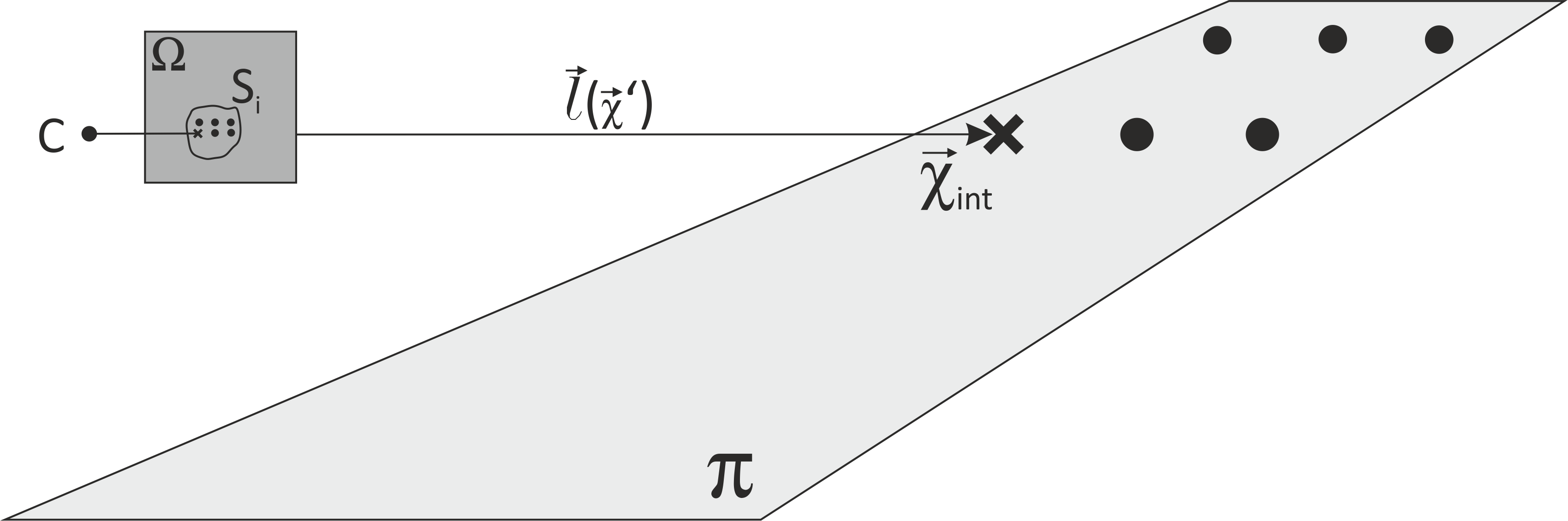}
		\caption{Interpolation method: Seeking the intersection of ray and plane}
		\label{fig:interpol}
	\end{center}
\end{figure}

That interpolation method is visualized in Figure~\ref{fig:interpol}. Using this method, all $\vec{x}^\prime_k$ can be interpolated. The interpolation method need to be restricted, because the approximated plane $\Pi$ might be near parallel to $\vec{l}(\vec{x}^\prime_k)$. Thus, there is a small intersection angle $\Theta$ which might introduce large outlier. Therefore, a restriction to $\Theta$ is introduced. The interpolation angle for each $\vec{x}^\prime_k$ is calculated using Equation~(\ref{eq:interpolAng}).

\begin{equation}\label{eq:interpolAng}
	\Theta(\vec{l}(\vec{x}^\prime_k), \vec{n}) = \arcsin \left( \frac{| \vec{n} \cdot \vec{l}(\vec{x}^\prime_k) |}{|| \vec{n} ||_2 \cdot || \vec{l}(\vec{x}^\prime_k) ||_2} \right)
\end{equation}

The restriction process is than done by applying Equation~(\ref{eq:interpolAngle}) to each $\vec{x}^\prime_k$

\begin{equation}
	\label{eq:interpolAngle}
	f_{int}(\Theta(\cdot)) = 
	\begin{cases} 0, & |\Theta(\cdot)| \leq \tau_{\Theta}\\
		Z_{int}, & else
	\end{cases}, \quad
	\forall \vec{x}^\prime_k \in U_i
\end{equation}

\noindent
which either assigns the interpolated depth $Z_{int}$ to $\vec{x}^\prime_k$ or leave it with unknown depth.

\subsection{Interpolation Loss}

\begin{figure}[ht]
	\begin{center}
		\includegraphics[width=0.48\textwidth]{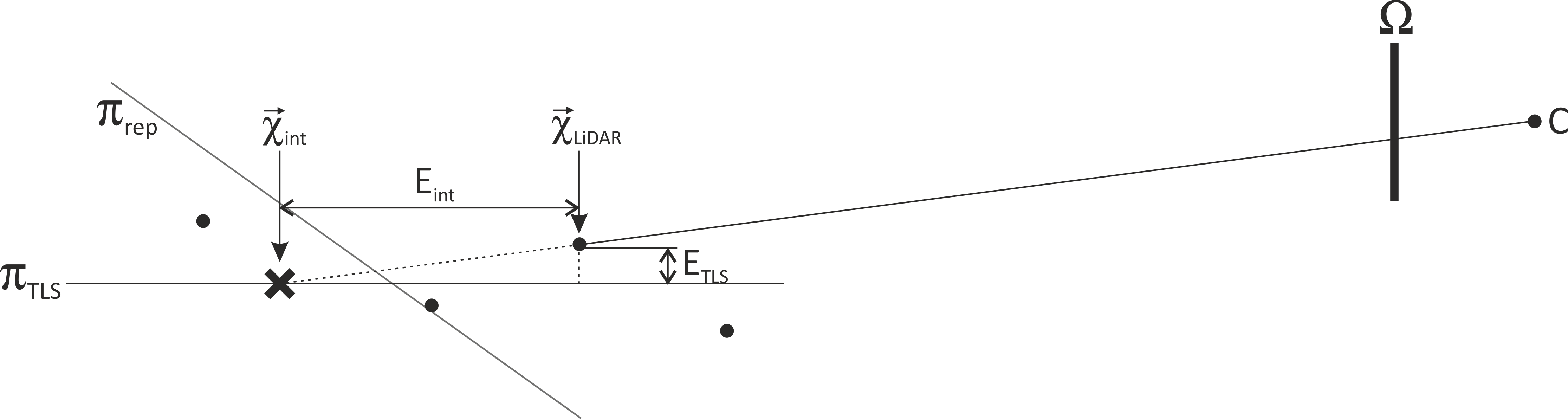}
		\caption{Interpolation loss vs orthogonal loss}
		\label{fig:eint_etls}
	\end{center}
\end{figure}

The interpolation loss $E_{int}$ is determined by equation~\ref{eq:reprojLoss1}. The difference of the interpolation error and the orthogonal loss is visualized in figure~\ref{fig:eint_etls}. It is visible, that the difference of the depth of a given~\gls{LiDAR} measurement and a lateral neighboring pixel (lying on the same plane) which need to be interpolated, can be very large. The smaller the intersection angle, the larger is the difference.

\begin{equation}
\label{eq:reprojLoss1}
	E_{int}(\vec{\beta}, \vec{l}(\vec{x}^\prime_k)) = \frac{1}{2} \sum_{k = 0}^{|K_i|} (Z_{lid, k} - Z_{int, k})^2
\end{equation}

with $\vec{\beta} = [\Pi_0, \; \Pi_1, \; \Pi_2, \; \Pi_3]^T$ and $\vec{l}(\vec{x}^\prime_k)) = [l_1, \; l_2, \; l_3]^T$ . $Z_{int}$ can be determined by equation \ref{eq:min}

\begin{equation}\label{eq:min}
	Z_{int} = w_3 + l_3 \cdot \frac{-\Pi_0 w_1 - \Pi_1 w_2 - \Pi_2 w_3}{\Pi_0 l_1 + \Pi_1 l_2 + \Pi_2 l_3} + p_{03}
\end{equation}

with a subscript denoting the $i$-th element of a vector and $p_{03}$ as the third element of $p_0$.

\subsection{Validity Check}
\label{sub:Validity}
After $\Pi$ is calculated, the boolean $B(\Pi)$ determines if $\Pi$ is valid (and thus will be used for interpolation) or not. $B(\Pi)$ is valid, if either the mean interpolation error is below a threshold $\tau_\Pi$ or the nearest point has a larger distance than $\tau_{far}$ while its error is below $\tau_{\Pi,far}$. This tolerates more error for distant points, which are more effected by the discretization error.

\begin{equation}
  \label{eq:val1}
  B(\Pi) =
  \begin{cases}
   \text{valid} &  \!\begin{aligned}%[b]
       & \frac{E_{int}(\cdot)}{|K_i|} \leq \tau_\Pi \; \lor \\
       & \; (min(Z_{int}) \geq \tau_{dist} \wedge \\
       & \frac{E_{int}(\cdot)}{|K_i|} \leq \tau_{\Pi,far})
    \end{aligned} \\%[1ex]
    \neg \text{valid} & \text{else}
  \end{cases}
\end{equation}

\subsection{Convex Hull Approach}

If the above mentioned plane is not valid due to too complex structure, the convex hull approach is employed which uses a modified RANSAC and aims to maximize the interpolated area. At first, multiple subsets of the superpixel corresponding pointcloud are used to determine a plane by solving equation \ref{eq:gauss} using Gaussian elimination

\begin{equation}
\label{eq:gauss}
	\begin{bmatrix}
		X_0 & Y_0 & 1 \\
		X_1 & Y_1 & 1 \\
		X_2 & Y_2 & 1\\
	\end{bmatrix}
	\begin{bmatrix}
		\Pi_0\\
		\Pi_1\\
		\Pi_3\\
	\end{bmatrix}
	=
	\begin{bmatrix}
		Z_0\\
		Z_1\\
		Z_2\\
	\end{bmatrix}
\end{equation}

with $\Pi_2 = -1$. Thus, multiple potential planes with a certain amount of inlier are obtained. The optimal loss to consider a point as inlier is the interpolation loss $E_{int}$ and an inlier is determined by the threshold $\tau_m$. Afterwards, the plane with the highest amount of inlier is sought which forms the vector $\vec{G} = \{ E_{int,1}, E_{int,2}, \dots, E_{int,p}\}$ which contains the interpolation error $E_{int}$ of all inliers $p$. Because usually $|S_i|$ is small, there exist multiple planes with the same amount of maximum inlier. Thus, the winning plane must satisfy

\begin{equation}	
	min \{ \rho(G_0), \rho(G_1), \dots, \rho(G_k) \}
\end{equation}

with 

\begin{equation}	
	\rho(G) = \frac{1}{|G|}\sum_{E_{int} \in G}E_{int}^2		
\end{equation}

The plane determined by the modified RANSAC is only used for interpolation, if it satisfies

\begin{equation}
	\label{eq:val2}
	B(\Pi) = 	\begin{cases} valid, & p \geq \tau_{abs} \lor \frac{p}{|K_i|} \geq \tau_{rel}\\		
	\neg valid, & else	\end{cases}
\end{equation}

If $B(\Pi)$ is valid, the plane is used to interpolate a convex hull, which is determined by the assigned image coordinates of the inlier $C = \left\{ (u_0,v_0), (u_1,v_1), \dots (u_p,v_p) \right\} \subset K_i$.

\subsection{Minimizing Interpolation Loss}
After $\Pi$ is determined, the interpolation loss is minimized (equation \ref{eq:interjLoss1}).

\begin{equation}
\label{eq:interjLoss1}
	\min_{\vec{\beta} \in \mathbb{R}^4} \frac{1}{2} \sum_{k = 0}^{|K_i|} (Z_{lid, k} - Z_{int, k})^2
\end{equation}

\begin{figure}[ht]
	\begin{center}
		\includegraphics[width=0.48\textwidth]{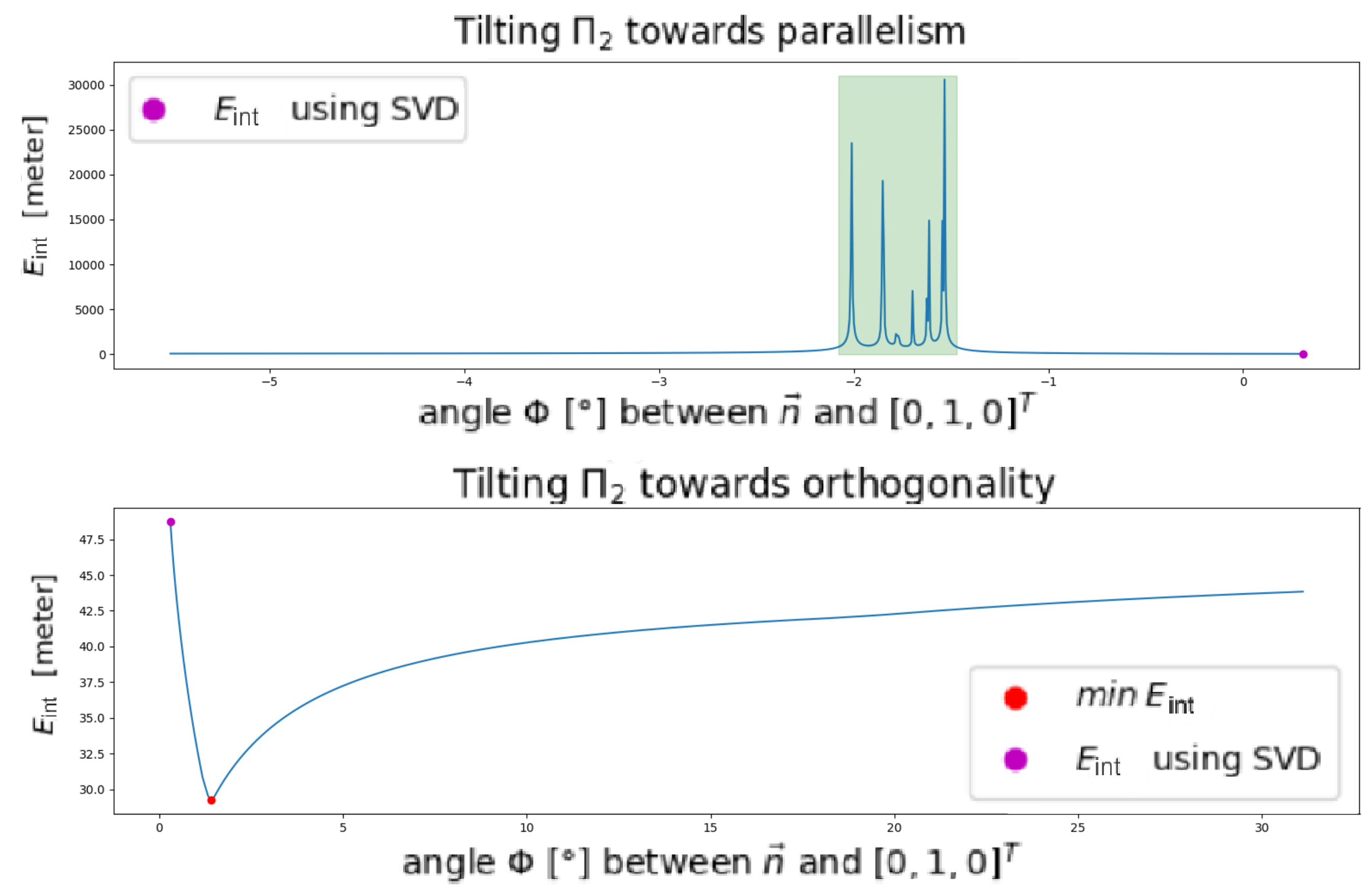}
		\caption{Interpolation error function. The more parallel the plane is, the larger the resulting error is.}
		\label{fig:obj}
	\end{center}
\end{figure}

This is done using the Truncated Newton algorithm and for runtime purposes using the analytical gradient information. Figure~\ref{fig:obj} visualizes the interpolation error of a ground plane. In the upper plot, the plane is tilted towards parallelism to the interpolation rays, while in the lower plot the plane is tilted towards orthogonality to the optical axis. It can be seen, minimizing the interpolation error tilts the planes towards to be parallel to the image plane / orthogonal to the optical axis. The starting point of each plot is determined using the singular value decomposition. 

\subsection{Interpolation of Remaining Areas}
\label{sub:rem}

\begin{figure}[ht]
	\begin{center}
		\includegraphics[width=0.48\textwidth]{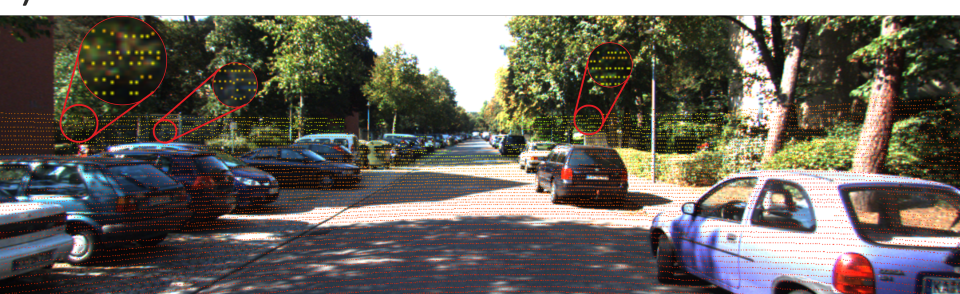}
		\caption{Poor image information and differing depths}
		\label{fig:poor}
	\end{center}
\end{figure}

Due to the convex hull approach in areas where the $S_i$ contains more than one object or too complex object structure, not interpolated areas remain. A typical scene with poor image information and differing depths in that area is shown in figure~\ref{fig:poor}. These areas need to be interpolated using other state-of-the-art deterministic approaches.

\subsection{Fusing Multiple Tentative Depth Maps}
To further increase the interpolated area, the image is oversegmented multiple times with different resolutions. This leads to different sized superpixel at different locations and therefore different interpolated areas. The final depth map is then obtained by fusing all tentative depth maps using Equation~(\ref{eq:fuse}).

\begin{multline}
\label{eq:fuse}
D_{fin}(u_k,v_k) = med(D_0(u_k,v_k), D_1(u_k,v_k),..., \\ D_n(u_k,v_k) ) \qquad , \forall (u_k,v_k) \in D 
\end{multline}

%%%%%%%%%%%%%%%%%%%%%%%%%%%%%%%%%%%%%%%%%%%%%%%%%%%%%%%%%%%%%%%%%%
\section{Experimental Work}
\label{sec:ExperimentalWork}
This section describes the used database, metrics, hardware and software setup and the used evaluation algorithms.

\subsection{Setup}
For a better reproducibility this section names the used hardware. The used CPU is a Intel(R) Core(TM) i7-8700 CPU @3.20GHz and 32 GB RAM. The code is written in Python and is vectorized.

\begin{table*}[ht]
	\centering
	\caption{Impact of different parameters in comparison to the setting: Colorspace = CIELAB, Number of depth maps = 1, Slic-iterations = 100, Optimizing interpolation loss = Yes, Convex hull approach = Yes, Fusion algorithm = IP-Basic}
	\label{tab:impact}
	\renewcommand{\arraystretch}{2.0}
	\begin{tabular}{l c c  c c c c} 
		%\hline 
		\specialrule{.3em}{.2em}{.2em}
		\textbf{Parameter} & \textbf{Altered value} & \textbf{MAE [mm]} & \textbf{RMSE [mm]} & \textbf{iMAE [1/km]} & \textbf{iRMSE [1/km]} & \textbf{Time [s]} \\ 
		%\hline 
		\specialrule{.3em}{.2em}{.2em}
		\specialrule{.3em}{.2em}{.2em}
		Number of depth maps & 3 & -3.389 & -0.625 & -0.026 & -0.067 & +5.799\\ 
		%\hline 
		%\hline 
		%SLIC-iter & 20 & -0.034 & -0.082 & +0.001 & +0.003 & -1.215\\ 
		\specialrule{.3em}{.2em}{.2em}
		\specialrule{.3em}{.2em}{.2em}
		%\hline 
		SLIC-iter& 5 & +0.166 & -0.818 & +0.002 & +0.003 & -1.445\\ 
		\hline 
		%	\hline 
		Colorspace& Gray & + 0.366 & -1.918 & +0.005 & +0.005 & -0.465\\ 
		\hline 
		%	\hline 
		SLIC-iter \& Colorspace& 20 \& Gray & +18.318 & +10.909 & +0.142 & +0.115 & -1.325\\ 
		%\hline 
		%\hline 
		\specialrule{.3em}{.2em}{.2em}
		\specialrule{.3em}{.2em}{.2em}
		Optimizing int. loss &  No & +1.166 & +0.782 & +0.013 & +0.016 & -0.535\\ 
		%\hline 
		%\hline 
		\specialrule{.3em}{.2em}{.2em}
		\specialrule{.3em}{.2em}{.2em}
		Convex Hull & No & +0.382 & +0.082 & +0.003 & +0.014 & -0.065\\ 
		%\hline 
		\specialrule{.3em}{.2em}{.2em}
	\end{tabular} 	
\end{table*}

\subsection{Metrics}

There are different metrics possible to analyze the proposed work based on different criteria. The first two metrics are the common Mean Absolute Error (MAE) as well as the~\gls{RMSE} (equation \ref{eq:RMSE}, \ref{eq:MAE}). Two more metrics are the inverse Mean Absolute Error (iMAE) and inverse Root Mean Squared Error (iRMSE) and are shown in equation \ref{eq:iRMSE}, \ref{eq:iMAE}. Both metrics calculate the error of the inverse depth. This results in a metric which gives errors occurring in small distances more weight than errors far away. 

Assume a ground truth depth map $G$ and a set of known ground truth points $N$:

\begin{equation}
\label{eq:RMSE}
	RMSE = 	\sqrt{\frac{1}{|N|} \sum_{k \in N} \left( G(k)-D(k)\right)^2}
\end{equation}

\begin{equation}
	\label{eq:MAE}
	MAE = \frac{1}{|N|} \sum_{k \in N}|G(k)-D(k)|
\end{equation}

\begin{equation}
	\label{eq:iRMSE}
	iRMSE = \sqrt{\frac{1}{|N|} \sum_{k \in N} \left( \frac{1}{G(k)}-\frac{1}{D(k)}\right)^2}
\end{equation}

\begin{equation}
	\label{eq:iMAE}
	iMAE = \frac{1}{|N|} \sum_{k \in N} \left| \frac{1}{G(k)}-\frac{1}{D(k)} \right|
\end{equation}

\subsection{Dataset}

There are multiple potential datasets such as the SYNTHIA~\cite{RosCVPR16}, Virtual KITTI~\cite{gaidon2016virtual, cabon2020vkitti2} and the KITTI Depth Completion benchmark~\cite{uhrig2017sparsity}. SYNTHIA and Virtual KITTI are synthetic datasets and thus have a very accurate ground truth but are not suitable for real-world applications, because they do not contain artifacts and a~\gls{LiDAR} sensor would need to be emulated. Therefore, the KITTI Depth Completion Benchmark is chosen as evaluation dataset. KITTI provides 93,000 training, 1,000 evaluation and 1,000 test images. The work is evaluated on the entire evaluation set.

%%%%%%%%%%%%%%%%%%%%%%%%%%%%%%%%%%%%%%%%%%%%%%%%%%%%%%%%%%%%%%%%%%

\section{Results and Discussion}
\label{sec:ResultsAndDiscussion}

In this section quantitative and qualitative results of the proposed paper are shown. The best non-deep learning based approach IP-Basic \cite{ku2018defense} (according to the KITTI ranking in terms of RMSE), the JBF~\cite{kopf2007joint} (transferred from ToF-superresolution) and two deep learning based approaches are used to evaluate the proposed approach. Remaining areas (section \ref{sub:rem}) are interpolated using the individual evaluation algorithm and is compared with its original version. For IP-Basic the Gaussian blur version and for Sparse-to-dense \cite{ma2019self} the deep regression model is used. % because thats the best setting.

At first, parameters are changed from a basic setting. This shows the impact (regarding errors and computational time) of different parameters such as the colorspace and employing the minimization of the interpolation loss. The basic setting (used for comparison) is named in the description of table \ref{tab:impact}, which shows the results:

\begin{itemize}
	
	\item At first, it shows that using more depth maps decrease the error slightly but increases the computational time significantly. 

	\item Furthermore, decreasing the SLIC iterations while using the CIELAB colorspace does have a huge impact while the error stays approximately constant. Although, reducing the SLIC iterations when using a grayscale image it reduces the computation time, but increases the error significantly and thus is not optimal. 

	\item Optimizing the interpolation loss for each plane slightly improves the error metrics but impacts the runtime significantly. 

	\item The convex hull does not have a huge impact on the runtime (due to its high vectorization), while it is slightly improving the error metrics. 

\end{itemize}

Thus, the best error runtime trade-off is achieved using CIELAB colorspace with 5 SLIC-iterations, without optimizing the interpolation loss and using the convex hull approach. This setting is called the final algorithm.

Finally, the deterministic joint bilateral filter~\cite{kopf2007joint} has been transferred from the ~\gls{ToF}-super-resolution field to the LiDAR depth completion field. Due to the irregular distribution of LiDAR points, a nearest neighbor approach replaces the fix-sized kernel.

\begin{table}[ht]
	\centering
	\caption{Error of final approach. Runtime fusion algorithm = IP-Basic is 0.504 seconds and when fusion algorithm = JBF runtime is 0.84 seconds.}
	\label{tab:finerr}
	\renewcommand{\arraystretch}{2.0}
	%\resizebox{0.45\textwidth}{!}{%
	\begin{tabular}{l c c c c c c}
		%\hline
		\specialrule{.3em}{.2em}{.2em}
		\textbf{Method} & \textbf{MAE} & \textbf{RMSE} & \textbf{iMAE} & \textbf{iRMSE} \\
		%\hline
		\specialrule{.3em}{.2em}{.2em}
		\specialrule{.3em}{.2em}{.2em}
		IP-Basic \cite{ku2018defense} & 305.352 & 1350.927 & 1.301 & 4.120\\
		\hline
		Final algorithm + \cite{ku2018defense} & \textbf{288.294} & \textbf{1339.971} & \textbf{1.173} & \textbf{4.021}\\
		%\hline
		%\hline
		\specialrule{.3em}{.2em}{.2em}
		\specialrule{.3em}{.2em}{.2em}
		JBF \cite{kopf2007joint} & 414.553 & 1609.884 & 1.786 & 6.197\\
		\hline
		Final algorithm + \cite{kopf2007joint} & \textbf{307.933} & \textbf{1296.610} & \textbf{1.248} & \textbf{3.921}\\
		%\hline
		%\hline
		\specialrule{.3em}{.2em}{.2em}
		\specialrule{.3em}{.2em}{.2em}
		Sparse-to-dense \cite{ma2019self} & 312.879 & \textbf{858.662} & 1.684 & 3.077\\
		\hline
		Final algorithm + \cite{ma2019self} & \textbf{262.838} & 878.686 & \textbf{1.167} & \textbf{2.767}\\
		%\hline
		%\hline
		\specialrule{.3em}{.2em}{.2em}
		\specialrule{.3em}{.2em}{.2em}
		deepLiDAR \cite{qiu2019deeplidar} & 215.136 & \textbf{687.001} & 1.094 & \textbf{2.505}\\
		\hline
		Final algorithm + \cite{qiu2019deeplidar} & \textbf{214.932} & 738.241 & \textbf{1.019} & 2.543 \\
		%\hline
		\specialrule{.3em}{.2em}{.2em}
	\end{tabular}%
	%}
\end{table}

Table \ref{tab:finerr} shows the error of the final algorithm. Bold values point out the approach with lower error, regarding the original evaluation algorithm and that approach used as fusion algorithm. It can be seen, that the proposed work outperforms the best non-deep learning based approach (IP-Basic - Rank 39 of all published and peer review paper in the KITTI depth completion benchmark) in all error metrics. Comparing IP-Basic to the JBF approach, the proposed work + JBF is significantly better regarding the iRMSE and RMSE and is the best combination (regarding RMSE) of deterministic approaches. Furthermore, this work enhances the MAE, iMAE and iRMSE of Sparse-to-dense (Rank 13) significantly. In comparison to deepLiDAR (Rank 6) the proposed work enhances the MAE and iMAE, while increasing the~\gls{RMSE} and iRMSE. 

Analyzing the metrics, it is seen that this work is especially good at interpolating near and large planes. However, it introduces outlier mainly far away (higher~\gls{RMSE} while iRMSE is moderate). This is caused by superpixel which violates object boundaries and can introduce large errors. Superpixel tend to undersegment objects far away due to the relation of object sizes and superpixel sizes, which is furthermore emphasized by worse color information for distant points.

\begin{figure}[ht]
	\begin{center}
		\includegraphics[width=0.48\textwidth]{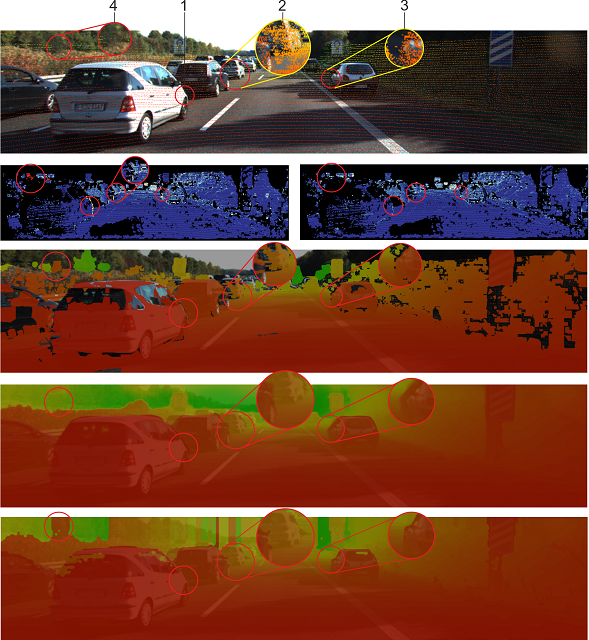}
		\caption{Qualitative results using multiple depth maps; Top-Figure: Projected LiDAR measurements (yellow circle contains ground truth); 2nd row: error maps of the proposed work (left) and evaluation algorithm (right); 3rd row: depth map of the proposed work (without interpolating remaining areas with evaluation algorithm); 4th row: Evaluation algorithm (deepLiDAR). Last row: IP-Basic. Area 1, 2, 3 show ground truth error. Area 4 shows better artifact handling of deepLiDAR.}
		\label{fig:res}
	\end{center}
\end{figure}

\begin{figure}[ht]
	\begin{center}
		\includegraphics[width=0.48\textwidth]{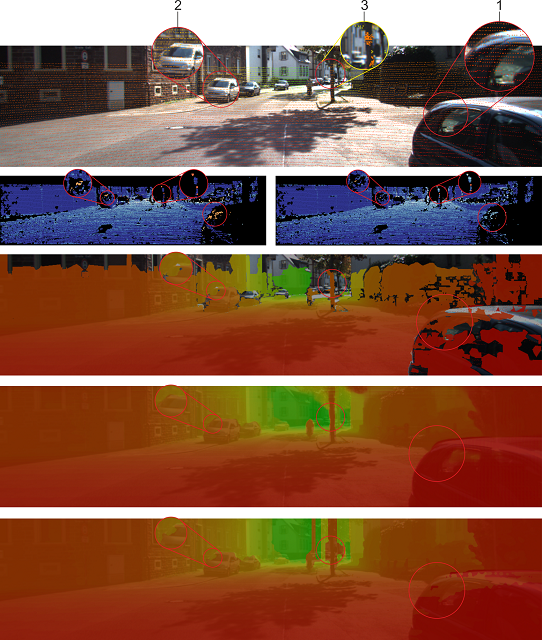}
		\caption{Qualitative results using multiple depth maps; Better artifact handling of deepLiDAR (4th row) in marked area 1. The proposed approach does not use global context and relies and the local measurements (groundplane measurements projected into the car, due to different acquisition times and the glass surface). Area 3 shows ground truth error. Area 2 shows object boundary violation.}
		\label{fig:art}
	\end{center}
\end{figure}

Figure~\ref{fig:res} shows a ground truth error which causes a higher~\gls{RMSE}. Marked circles point out areas to focus on. Area 1, 2 (black and white car), 3 shows a weakness of the ground truth and explains the constant higher~\gls{RMSE}. Looking at the ground truth points it can be seen that the depths are not perfectly aligned with the image information. Although the proposed algorithm is perfectly aligned with the object boundaries, the ground truth error maps show large outlier (and small outlier for deepLiDAR although it is clearly violating the boundaries). This artifact occurs frequently at front bumpers. Finally, a higher~\gls{RMSE} is also caused by the better artifact handling of deep learning based approaches which can be seen in Figure~\ref{fig:art}, where deep learning based are able to use more global information to approximate the car (area 1).

%\begin{figure}[ht]
%	\begin{center}
%		\includegraphics[width=0.48\textwidth]{3}
%		\caption{depth map when using only 1 segmented image}
%		\label{fig:fin1}
%	\end{center}
%\end{figure}

%%%%%%%%%%%%%%%%%%%%%%%%%%%%%%%%%%%%%%%%%%%%%%%%%%%%%%%%%%%%%%%%%%

\section{Conclusion and Future Recommendations}
\label{sec:conclusion}

This paper proposed an algorithm which fuses camera and LiDAR data in a deterministic way, to densify a LiDAR based depth map. The image is divided into multiple superpixel and the pinhole camera model is used for the interpolation of each superpixel represented by a plane. This work outperforms the best non-deep learning based approaches and several deep learning based approaches in the KITTI depth completion challenge.

Because each superpixel is processed sequentially, future work should focus on the parallelization. Furthermore, future algorithms should tackle the removal of artifacts as well as how to approximate non-planar areas. 

%%%%%%%%%%%%%%%%%%%%%%%%%%%%%%%%%%%%%%%%%%%%%%%%%%%%%%%%%%%%%%%%%%
\vfill
%\section*{ACKNOWLEDGMENT}

%%%%%%%%%%%%%%%%%%%%%%%%%%%%%%%%%%%%%%%%%%%%%%%%%%%%%%%%%%%%%%%%%%
%\addtolength{\textheight}{-12cm}
%\vspace{10mm}
\bibliographystyle{IEEEtran}
\bibliography{paper}
\end{document}